\documentclass[runningheads]{llncs}
\usepackage{graphicx}

\usepackage{tikz}
\usepackage{comment}
\usepackage{amsmath,amssymb} 
\usepackage{color}
\usepackage{url}
\usepackage[accsupp]{axessibility} 

\usepackage{here}
\usepackage{caption}
\usepackage{multirow}
\usepackage{chngpage}
\usepackage{arydshln}
\usepackage{gensymb}
\usepackage{url}

\usepackage{xspace}
\def\eg{\emph{e.g.,}\xspace}
\def\ie{\emph{i.e.,}\xspace}

\def\etal{\emph{et al.}\xspace}

\begin{document}

\pagestyle{headings}
\mainmatter

\title{UnrealEgo: A New Dataset for Robust Egocentric 3D Human Motion Capture}

\titlerunning{UnrealEgo: A New Dataset for Robust Egocentric 3D Human MoCap}

\author{Hiroyasu Akada\inst{1,2}
\and
Jian Wang\inst{1} 
\and
Soshi Shimada\inst{1} \and
Masaki Takahashi\inst{2} \and \\
Christian Theobalt\inst{1} \and
Vladislav Golyanik\inst{1}
}
\authorrunning{H. Akada et al.}
\institute{
Max Planck Institute for Informatics, SIC 
\and
Keio University 
}

\maketitle

\begin{abstract}
We present \textit{UnrealEgo}, \ie a new large-scale naturalistic dataset for egocentric 3D human pose estimation.
UnrealEgo is based on an advanced concept of eyeglasses equipped with two fisheye cameras that can be used in unconstrained environments. 
We design their virtual prototype and attach them to 3D human models for stereo view capture. 
We next generate a large corpus of human motions. 
As a consequence, UnrealEgo is the first dataset to provide in-the-wild stereo images with the largest variety of motions among existing egocentric datasets. 
Furthermore, we propose a new benchmark method with a simple but effective idea of devising a 2D keypoint estimation module for stereo inputs to improve 3D human pose estimation. 
The extensive experiments show that our approach outperforms the previous state-of-the-art methods qualitatively and quantitatively. 
UnrealEgo and our source codes are available on our project web page\footnote{\url{https://4dqv.mpi-inf.mpg.de/UnrealEgo/}}.

\end{abstract}

\begin{figure*}[t]
  \centering
  \includegraphics[height=4.23cm]{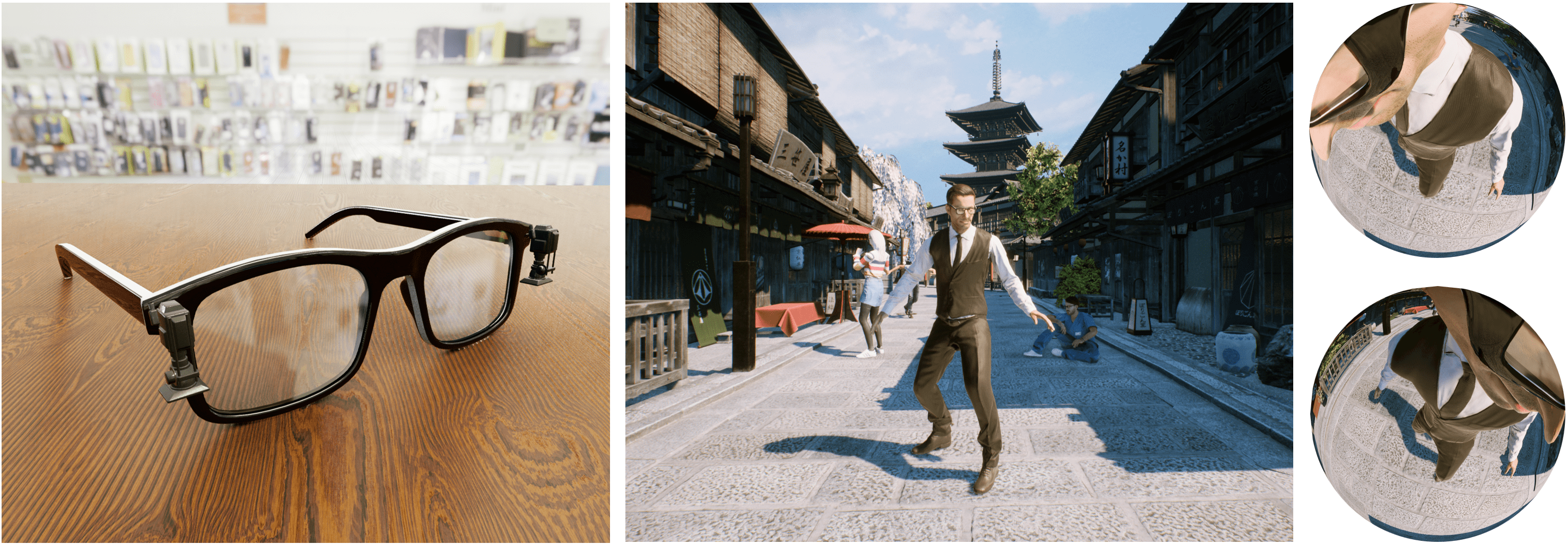}\\
    \caption*{\scriptsize
        (a) {\begin{tabular}[c]{@{}c@{}} Proposed concept of glasses \\equipped with two fisheye cameras\end{tabular}} 
        \,\,\,\,\,\,\,\,\,\,\,\,\,\,\,\,\,\,\,\,\,\,\,
        (b) {\begin{tabular}[c]{@{}c@{}} Human model \\wearing the glasses\end{tabular}} 
        \,\,\,\,\,\,\,\,\,\,\,\,\,\,\,\,\,\,\,\,\,\,\,\,\,\,
        (c) {\begin{tabular}[c]{@{}c@{}} Egocentric \\fisheye views\end{tabular}} 
        \,\,\,\,\,\,
    }
  \caption{Overview of the proposed UnrealEgo setup.} 
  \label{fig:overview} 
\end{figure*}

\section{Introduction}
\label{introduction}

Egocentric 3D human pose estimation has been actively researched recently \cite{rhodin2016egocap,xu2019mo2cap2,jiang2021egocentric,tome2019xr,Tom2020SelfPose3E,zhang2021automatic,wang2021estimating,zhao2021egoglass}. 
Compared to cumbersome motion capture systems that require a fixed recording volume, the egocentric setup is more suitable to capture daily human activities in unconstrained environments. 
Example applications include XR technologies \cite{ElgharibMendiratta2020} and 
motion analysis for sport and health \cite{NunezMarcos2022}.

Several setup types were proposed for egocentric 3D human pose estimation. 
Some methods work on mobile devices such as a cap~\cite{xu2019mo2cap2}, a helmet~\cite{rhodin2016egocap} or a head-mounted display~\cite{tome2019xr,Tom2020SelfPose3E} equipped with a camera to capture egocentric views of a user's whole body. 
Although these methods show promising results, their setups are still not satisfactory for daily use; 
the cameras are mounted far from the user's body, which is inconvenient and restrictive. 
The recently introduced EgoGlass  approach~\cite{zhao2021egoglass} tackles this  issue by an eyeglasses-based setup with two cameras attached to the glasses frame. 
Their setup imposes fewer restrictions on users' activities. 
We envision that with the recent development of smaller cameras \cite{calicam} and smart glasses~\cite{googleglasses,rayban}, the eyeglasses-based setup can be a de facto standard to capture daily human activities in various situations. 

Along with that, there is a lack of datasets that would account for this new and advanced capture setting and  that would allow developing algorithmic frameworks involving it. 
Furthermore, existing egocentric datasets are limited in several ways and cannot be easily re-purposed for 3D human pose estimation with the compact eyeglasses-based setup. 
First, the existing datasets do not contain complex human motions (such as breakdance and backflip) that are seen in daily human activities~\cite{rhodin2016egocap,xu2019mo2cap2,tome2019xr,zhao2021egoglass}. 
Second, the available egocentric datasets do not faithfully model the 3D environment~\cite{xu2019mo2cap2,tome2019xr}. 
Next, the existing stereo-based datasets~\cite{rhodin2016egocap,zhao2021egoglass} do not contain in-the-wild images. 
All in all, we note that there is no large-scale stereo-based dataset currently available. 
Consequently, a lack of a comprehensive and versatile egocentric dataset is a severely limiting factor in the development of methods for egocentric 3D perception. 

\begin{figure}[t]
 \centering
 \includegraphics[width=12cm]{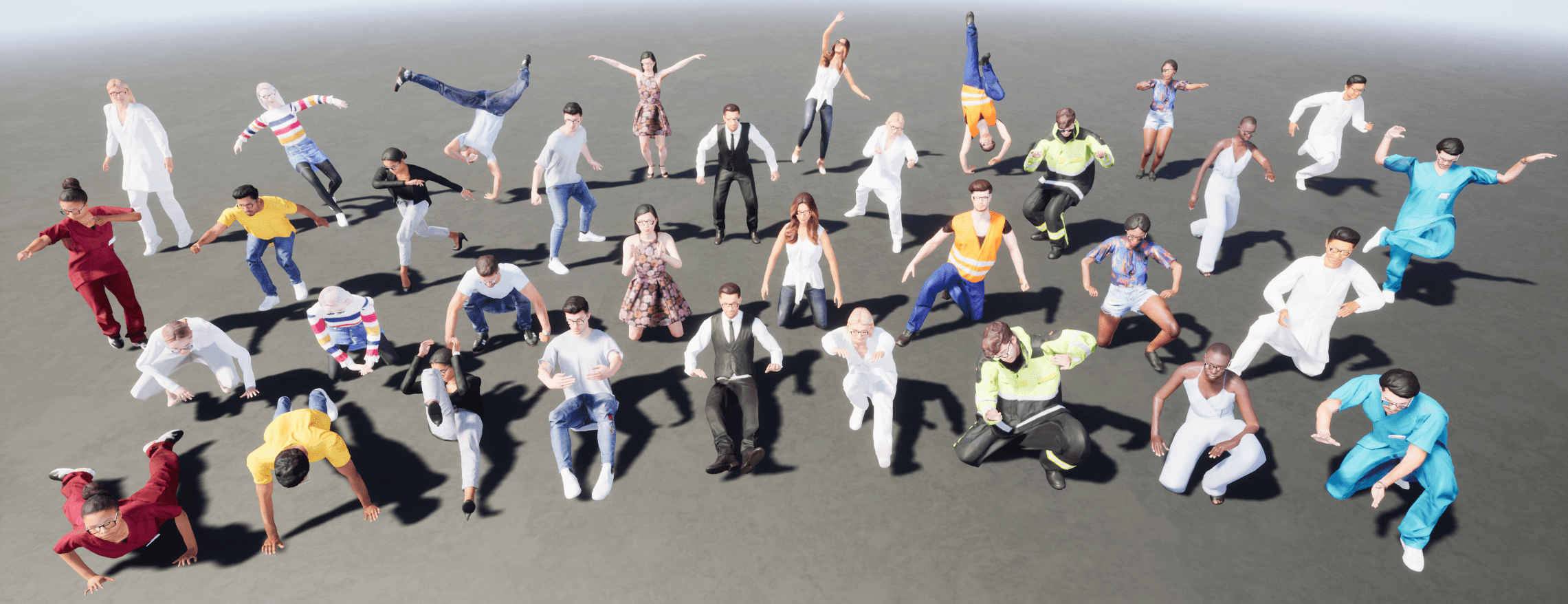}
 \caption{Samples of characters and poses from UnrealEgo. 
 We use 17 high-quality 3D RenderPeople models ~\cite{renderpeople}. 
 Also, we utilize Mixamo motions~\cite{mixamo} and modify them to diversify our motion data.
 Please refer to our video for better visualizations and our supplementary asset list for characteristics of each human model. 
 }
 \label{fig:characters}
\end{figure}

To alleviate the issues mentioned above, we present \textit{UnrealEgo}, \ie a new large-scale naturalistic and synthetic dataset for egocentric 3D human pose estimation. 
UnrealEgo is based on an advanced concept of an eyeglasses-based  setup with two fisheye cameras symmetrically attached to the glasses frame. 
Fisheye cameras are getting more and more compact; they can capture a wider range of views than normal cameras which is beneficial for egocentric human pose estimation~\cite{rhodin2016egocap}. 
We use Unreal Engine~\cite{unrealengine} to synthetically design the eyeglasses as shown in Fig.~\ref{fig:overview}-(a). 
We then attach the eyeglasses to realistic 3D human models (\textit{RenderPeople})~\cite{renderpeople} and capture in-the-wild stereo views in various 3D environments as shown in Fig.~\ref{fig:overview}-(b), (c). 
Note that we prioritize the motion diversity in UnrealEgo. 
Fig.~\ref{fig:characters} shows examples of 3D human models in diverse poses from UnrealEgo. 
In total, UnrealEgo contains $450$k in-the-wild stereo views ($900$k images in total) with the largest variety of motions among the existing egocentric datasets. 
UnrealEgo allows developing new methods that account for temporal changes of surrounding 3D environments (see Sec.~\ref{unrealego}) and evaluating the current state-of-the-art methods in highly challenging scenarios (see Sec.~\ref{experiments}). 

Furthermore, we propose a new benchmark approach that achieves state-of-the-art accuracy on UnrealEgo. 
At the core of our method is a heatmap-based 2D keypoint estimation module. 
It accepts stereo inputs and passes them to two weight-sharing encoders that produce feature maps in the latent space. 
The obtained feature maps are concatenated along with the channel dimensions and processed by a decoder that estimates 2D keypoint heatmaps (see Fig.~\ref{fig:overview of method}). 
In extensive experiments, we observe that this simple but effective architecture brings significant improvements compared with existing methods~\cite{tome2019xr,zhao2021egoglass} qualitatively and quantitatively by 13.5\% on MPJPE and 14.65\% on PA-MPJPE metrics. 

In summary, the primary \textbf{contributions} of this work are as follows:  
\begin{itemize} 
    \item \textit{UnrealEgo}, \ie a new large-scale naturalistic dataset for egocentric 3D human motion capture. 
    \item A new approach for 3D human pose estimation achieving state-of-the-art accuracy on the new benchmark dataset. 
\end{itemize} 

UnrealEgo is the first to provide 1) naturalistic in-the-wild stereo images with the largest variety of motions and 2) sequences with realistically and accurately-modeled changes of the surrounding 3D environments.
This allows a more thorough evaluation of existing and upcoming methods for egocentric 3D vision, including the temporal component and global 3D poses. 

\section{Related Work}

\subsection{Datasets for Outside-in 3D Human Pose Estimation} 
Many datasets were proposed for 3D pose estimation with ground-truth annotations. 
Some of them are captured with optical markers~\cite{HEva_Sigal:IJCV:10b,h36m_pami,Trumble:BMVC:2017}, while the others use marker-less mocap  systems~\cite{mehta2018single,mono-3dhp2017,7410738,yu2020humbi}. 
However, these datasets are mostly captured in the studio and usually lack the diversity of clothing, occlusions, and environments.

In the meantime, synthetic datasets have become popular because no costly mocap setups are required for annotations. 
Many such datasets are created by compositing people on background images~\cite{Varol:CVPR:2017,pumarola20193dpeople,Hoffmann:GCPR:2019,Ranjan:IJCV:2020,mono-3dhp2017,singleshotmultiperson2018}. 
Because of such composition, however, their images do not match real-world scenes in terms of the local pixel intensity statistics and distributions. 
Butler \etal~\cite{Butler:ECCV:2012} provide images rendered using underlying detailed 3D geometry and corresponding optical flows that can be used for tracking purposes. However, this dataset does not provide 3D joint annotations unlike ours.

The recent works by Zhu~\etal~\cite{zhu2020simpose} and Patel~\etal~\cite{Patel:CVPR:2021} use 3D modeling tools and game engines~\cite{blender,unity,unrealengine} to render realistic images of rigged 3D human models in 3D environments. 
Unfortunately, these datasets are designed for outside-in pose estimation from an external camera viewpoint; they are not suitable and cannot be easily repurposed for egocentric 3D pose estimation. 

\subsection{Datasets for Egocentric 3D Human Pose Estimation} 
\label{related work:egocentric data}
There exist several datasets specifically recorded for egocentric 3D human poses. 
Mo$^2$Cap$^2$~\cite{xu2019mo2cap2} is the first cap-based setup with a single wide-view fisheye camera attached 8cm away from the user. %
With this setup, Xu~\etal~\cite{xu2019mo2cap2} 
create a large-scale dataset by compositing SMPL models~\cite{SMPL:2015} on randomly-chosen  backgrounds 
(real images), resulting in $530$k images with 15 annotated keypoints per image. 
xR-EgoPose~\cite{tome2019xr} approach uses a head-mounted display with a single fisheye camera equipped 2cm away from a user's nose. 
This work uses the Mixamo motion dataset~\cite{mixamo} to animate 3D human models and renders egocentric views with HDR backgrounds with the help of the 3D rendering tool V-Ray~\cite{vray}. 
Their dataset contains $380$k photorealistic synthetic images with $25$ body and $40$ hand keypoints. 
However, both datasets contain only monocular images. 
They feature only simple (every-day) human motions (due to the restrictions imposed by their setups) and do not accurately model 3D environments and complex human trajectories in them. 
Hence, they do not cover most motions that can arise in egocentric 3D human pose estimation using a compact eyeglass-based setup. 
Ego4D \cite{Grauman_etal_2022_CVPR} is a new large-scale dataset for egocentric vision. 
Unfortunately, it does not contain 3D annotations of human poses.

On the other hand, existing stereo egocentric datasets have several limitations. 
Rhodin \etal~\cite{rhodin2016egocap} proposed EgoCap, \ie a headgear with a pair of fisheye cameras equipped 25cm away from users to capture stereo views. 
Their dataset contains only 30k stereo image pairs with a limited variety of motions in a lab environment. 
More recently, EgoGlass~\cite{zhao2021egoglass} simplified the stereo setup with eyeglasses and two cameras equipped on the glasses frames. 
Although EgoGlass captured a relatively large-scale of images, \ie total 170k stereo pairs, 
\textit{the dataset is captured only in a studio environment and is not publicly available.}  

\begin{figure}[t]
 \centering
 \includegraphics[width=12cm]{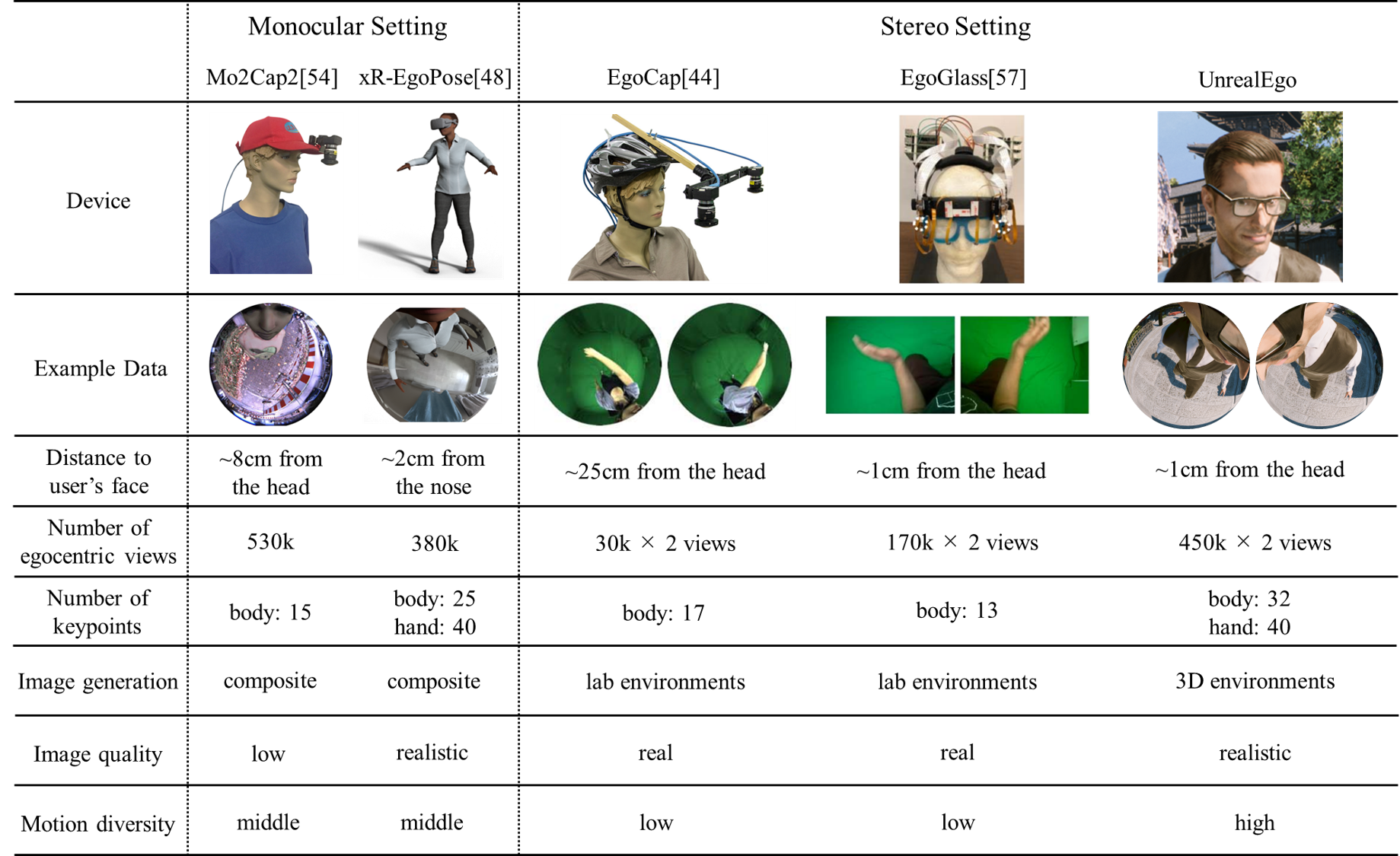}
 \caption{Comparison of datasets for egocentric 3D human pose estimation.}
 \label{fig:egocap}
\end{figure}

In contrast to existing datasets, UnrealEgo addresses the above shortcomings. 
Fig.~\ref{fig:egocap} illustrates the differences among existing datasets and UnrealEgo. 
Firstly, UnrealEgo provides stereo images in indoor and outdoor scenes.
Secondly, it offers the largest number of images, \eg 15 times larger than EgoCap \cite{rhodin2016egocap} and 2.5 times larger than EgoGlass \cite{zhao2021egoglass}.
Next, it contains naturalistic image sequences with accurately  modeled geometry changes in the surrounding 3D environments. 
Also, it offers the largest number of body and hand keypoints.
Furthermore, it is the most challenging egocentric dataset in terms of motion variety.

\subsection{Methods for Egocentric 3D Human Pose Estimation} 
Existing methods for egocentric 3D human pose estimation can be divided into two groups in terms of egocentric settings. 
The first group aims at estimating 3D keypoints from monocular views. 
Mo2Cap2~\cite{xu2019mo2cap2} is the first CNN-based system to predict 3D poses.
Tome \etal~\cite{tome2019xr,Tom2020SelfPose3E} follow a two-step approach using a multi-branch autoencoder to capture uncertainty in their predicted 2D heatmaps and to leverage rotation  constraints~\cite{Tom2020SelfPose3E}. 
Jiang \etal~\cite{jiang2021egocentric} predict 3D poses by utilizing the information of surrounding environments and extremities of the user's body. 
Zhang \etal~\cite{zhang2021automatic} estimate 3D poses with 
fisheye distortions using an automatic calibration module. 
More recently, Wang \etal~\cite{wang2021estimating} proposed an optimization-based approach with a motion prior learned from an additional dataset for global 3D human motion capture. 
Even with their competitive results, these monocular methods often fail on complex motions (\eg due to the depth ambiguity). 

The second group follows multi-view settings, including our work. 
EgoCap~\cite{rhodin2016egocap} is an optimization-based approach using a body-part detector and personalized 3D skeleton models. 
Cha \etal~\cite{8458443} developed a headset 
equipped
with eight cameras; they introduced a CNN-based method to  reconstruct a human body and an environment in 3D. 
EgoGlass~\cite{zhao2021egoglass} builds upon xR-EgoPose~\cite{tome2019xr} and is one of the most accurate methods; 
its architecture contains two separate UNets for the stereo inputs in the 2D joint estimation module. 
In contrast to the reviewed works, this paper proposes a simple yet effective idea of devising a new 2D joint estimation module that accepts stereo inputs to significantly improve 3D pose estimation compared with the existing best-performing methods.

\begin{table}[t]
\begin{center}
\caption{Comparison of human motion capture datasets. 
}
\label{table:comparison_mocap_datasets}
\scalebox{0.85}{
\begin{tabular}{lrrr|lrrr}
\noalign{\smallskip}
\hline
Dataset                                & Subjects & Motions & Minutes \,\,&\,\, Dataset & Subjects & Motions & Minutes   \,  \\
\hline
ACCAD \cite{ACCAD}                             & 20       & 252     & 26.74    \,\,&\,\, KIT \cite{KIT_Dataset}                         & 55       & 4232    & 661.84   \,  \\
BMLhandball \cite{BMLhandball}                 & 10       & 649     & 101.98   \,\,&\,\, MPI HDM05 \cite{MPI_HDM05}                     & 4        & 215     & 144.54   \,  \\
BMLmovi \cite{ghorbani2020movi}                & 89       & 1864    & 174.39   \,\,&\,\, MPI Limits \cite{PosePrior_Akhter:CVPR:2015}   & 3        & 35      & 20.82    \,  \\
BMLrub \cite{BMLrub}                           & 111      & 3061    & 522.69   \,\,&\,\, MPI MoSh \cite{MoSh_lopermahmoodetal2014}      & 19       & 77      & 16.53    \,  \\
CMU \cite{cmuWEB}                              & 96       & 1983    & 543.49   \,\,&\,\, MPI-INF-3DHP \cite{mono-3dhp2017}              & 8        & -       & -        \,  \\
D-FAUST \cite{dfaust:CVPR:2017}                & 10       & 129     & 5.73     \,\,&\,\, SFU \cite{SFU}                                 & 7        & 44      & 15.23    \,  \\
DanceDB \cite{DanceDB:Aristidou:2019}          & 20       & 151     & 203.38   \,\,&\,\, SSM \cite{AMASS:2019}                          & 3        & 30      & 1.87     \,  \\
EKUT \cite{KIT_Dataset}                        & 4        & 349     & 30.74    \,\,&\,\, TCD Hands \cite{TCD_hands}                     & 1        & 62      & 8.05     \,  \\
Eyes Japan \cite{Eyes_Japan}                   & 12       & 750     & 363.64   \,\,&\,\, TotalCapture \cite{Trumble:BMVC:2017}          & 5        & 37      & 41.1     \,  \\
Human3D \cite{h36m_pami}                       & 11       & -       & -        \,\,&\,\, Transitions \cite{AMASS:2019}                  & 1        & 110     & 15.1     \,  \\
Human4D \cite{chatzitofis2020human4d}          & 8        & 148     & 72.60    \,\,&\,\, AMASS \cite{AMASS:2019}                        & 344      & 11265   & 2420.86  \,  \\ \cdashline{5-8} 
HumanEva \cite{HEva_Sigal:IJCV:10b}            & 3        & 28      & 8.48     \,\,&\,\, Ours                                           & 17       & 45520   & 3174.63  \,  \\ 
\hline
\end{tabular}
}
\end{center}
\end{table}

\section{UnrealEgo Dataset} 
\label{unrealego} 
This section provides details of the UnrealEgo dataset, focusing on our setup, motions, and rendered egocentric data. 
Please also see our supplementary video for dynamic visualizations and our supplementary asset list.

\subsection{Setup}
\label{setup}
We use Unreal Engine \cite{unrealengine} to synthetically design the eyeglasses with two fisheye cameras equipped on the glasses frame as shown in  Fig.~\ref{fig:overview}-(a). 
The distance between the cameras is 12$\mathrm{cm}$. 
The cameras' field of view amounts to $170\degree$.
We attach the glasses to 3D human models (RenderPeople) that perform different motions in various 3D environments. 
Fig.~\ref{fig:overview}-(b) and (c) show an example of the human models in a Kyoto-inspired environment in Japan, and fisheye views. 

\smallskip
\noindent\textbf{Characters.} We use 17 realistic RenderPeople 3D human models (commercially available) \cite{renderpeople}, nine female and eight male. These models are rigged and skinned based on the default 3D human skeleton of Unreal Engine~\cite{unrealengine}. 
Their skin color tones include pale white, white, light brown, moderate brown, dark brown, and black. 
Their clothing types include athletic pants, jeans, shorts, tights, dress pants, skirts, jackets, t-shirts, and long sleeves with diffident colors. 
Please see Fig.~\ref{fig:characters} for an overview of the 3D human models we use. %
Also, please see our supplement for detailed  characteristics of each human model.

\begin{table}[t]
\begin{center}
\caption{Motion categories in our dataset.} 
\label{table:motion_category_unrealego}
\scalebox{0.87}{
\begin{tabular}{lrr|lrr}
\hline
Motion types                                           & Motions  & Minutes     \,\,&\,\, Motion types                                & Motions  & Minutes     \,  \\
\hline
\,\,\,\,\,\,\,\,1:\,\,\,jumping                        & 1343       & 36.35     \,\,&\,\,16:\,\,\,standing - whole body               & 3791       & 307.95    \,  \\
\,\,\,\,\,\,\,\,2:\,\,\,falling down                   & 714        & 35.27     \,\,&\,\,17:\,\,\,standing - upper body               & 5820       & 708.74    \,  \\
\,\,\,\,\,\,\,\,3:\,\,\,exercising                     & 1225       & 82.07     \,\,&\,\,18:\,\,\,standing - turning                  & 1785       & 82.73     \,  \\
\,\,\,\,\,\,\,\,4:\,\,\,pulling                        & 272        & 28.31     \,\,&\,\,19:\,\,\,standing - to crouching             & 680        & 38.21     \,  \\
\,\,\,\,\,\,\,\,5:\,\,\,singing                        & 1054       & 149.21    \,\,&\,\,20:\,\,\,standing - forward                  & 3417       & 93.68     \,  \\
\,\,\,\,\,\,\,\,6:\,\,\,rolling                        & 136        & 4.69      \,\,&\,\,21:\,\,\,standing - backward                 & 1207       & 21.69     \,  \\
\,\,\,\,\,\,\,\,7:\,\,\,crawling                       & 612        & 22.47     \,\,&\,\,22:\,\,\,standing - sideways                 & 1496       & 30.42     \,  \\
\,\,\,\,\,\,\,\,8:\,\,\,laying                         & 612        & 30.92     \,\,&\,\,23:\,\,\,dancing                             & 5728       & 800.13    \,  \\
\,\,\,\,\,\,\,\,9:\,\,\,sitting on the ground          & 68         & 10.88     \,\,&\,\,24:\,\,\,boxing                              & 4012       & 160.53    \,  \\
\,\,\,\,\,10:\,\,\,crouching - normal                  & 1802       & 127.90    \,\,&\,\,25:\,\,\,wrestling                           & 2958       & 119.63    \,  \\
\,\,\,\,\,11:\,\,\,crouching - turning                 & 612        & 12.74     \,\,&\,\,26:\,\,\,soccer                              & 1892       & 69.63     \,  \\
\,\,\,\,\,12:\,\,\,crouching - to standing             & 850        & 29.46     \,\,&\,\,27:\,\,\,baseball                            & 476        & 27.31     \,  \\
\,\,\,\,\,13:\,\,\,crouching - forward                 & 1020       & 29.50     \,\,&\,\,28:\,\,\,basketball                          & 272        & 7.54      \,  \\
\,\,\,\,\,14:\,\,\,crouching - backward                & 493        & 8.82      \,\,&\,\,29:\,\,\,american football                   & 85         & 6.07      \,  \\
\,\,\,\,\,15:\,\,\,crouching - sideways                & 646        & 11.69     \,\,&\,\,30:\,\,\,golf                                & 442        & 80.07     \,  \\

\hline
\end{tabular}
}
\end{center}
\end{table}

\smallskip
\noindent\textbf{Motions.} 
It is our top priority to include a wider variety of motions that can represent as many daily human activities as possible. Therefore, we first create a new large corpus of motions. 
Specifically, we utilize Mixamo motions~\cite{mixamo} and modify them using Unreal Engine \cite{unrealengine} to enhance their plausibility and diversify the motion data. 
We first manually fix some motions that involve self-penetration and then modify the motions in various ways, including the speed of motions, arm movements, foot stances, and head rotations. 
For further details, please refer to our supplement.
In total, we created 45,520 natural motions for the 17 human models, \textit{i.e.,} ${\approx}2700$ motions per model.
We provide the details of our dataset %
in Tables~\ref{table:comparison_mocap_datasets} and \ref{table:motion_category_unrealego}. 
Table~\ref{table:comparison_mocap_datasets} compares existing mocap datasets and our motion data. 
Note that AMASS \cite{AMASS:2019} is a collection of several existing motion capture datasets~\cite{ACCAD,BMLhandball,BMLrub,ghorbani2020movi,cmuWEB,dfaust:CVPR:2017,DanceDB:Aristidou:2019,KIT_Dataset,Eyes_Japan,chatzitofis2020human4d,HEva_Sigal:IJCV:10b,KIT_Dataset,MPI_HDM05,PosePrior_Akhter:CVPR:2015,MoSh_lopermahmoodetal2014,SFU,TCD_hands,Trumble:BMVC:2017}. 
Our dataset contains the largest number of motions with the longest 
consecutive 3D human motions. 
Table~\ref{table:motion_category_unrealego} summarises the included motion categories.

\begin{figure}[t]
 \centering
 \includegraphics[width=12cm]{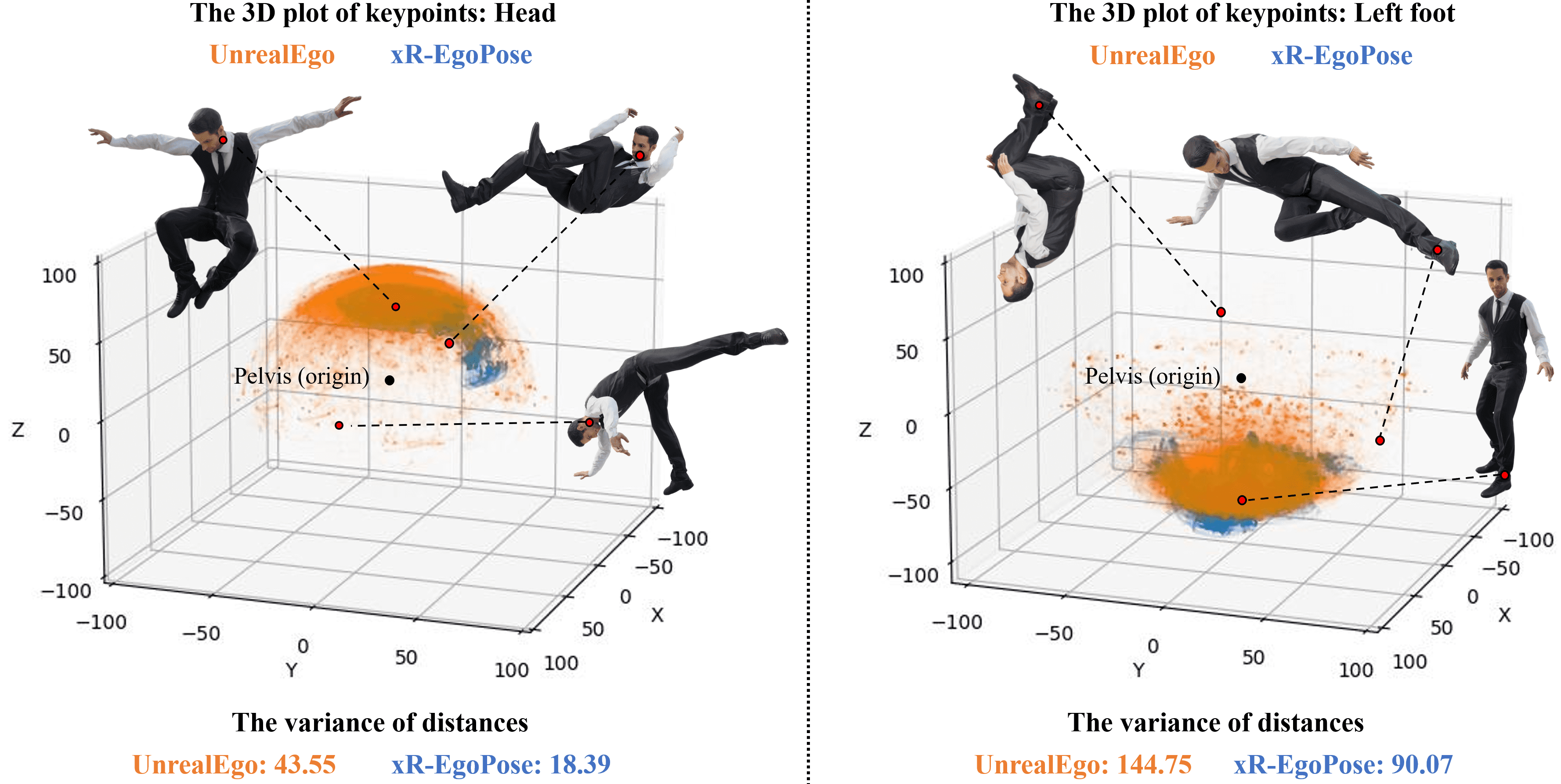}
 \caption{Distributions of head and left foot locations in  xR-EgoPose~\cite{tome2019xr} (blue) and UnrealEgo (orange). 
 The pelvis-relative 3D coordinates are on $\mathrm{cm}$-scale. 
 } 
 \label{fig:distribution of keypoints}
\end{figure}

\smallskip
\noindent\textbf{3D Environments.} 
We use 14 realistic 3D environments. They include a variety of indoor and outdoor scenes (\eg parks, roads, bridges, offices, gardens, playrooms, laboratories, cafeterias, trains, tennis courts, baseball fields, football fields, factories, European boulevards, North-American houses, Chinese rooms, Kyoto towns, and Japanese restaurants, at different times of day and night). Please see our supplementary asset list for further details. 

\smallskip
\noindent\textbf{Spawning Human Characters.}
It is important to create populated scenes to simulate real-world situations. 
To this end, we develop an algorithm to randomly place human models in 3D environments in Unreal Engine. 
As a preliminary step, we manually place $K$
rectangles 
$B{=}\,\{B_{1},...,B_{K}\}$ where several human models can be spawned on even grounds. 
Here, let $S{=}\,\{S_{1},...,S_{K}\}$ be the areas of 
rectangles 
and $C{=}\,\{C_{1},...,C_{K}\}$ be their center positions in the  world frame in Cartesian coordinate, respectively. 
As a first step, we choose $i$-th area $B_{i}\in B$ using area weighted probability $S_{i}/\sum_{i=1}^{K}S_{i}$. 
Secondly, we select $T$ surrounding rectangles $B_{t} \in B$, $t = \{1, \ldots, T\}$ with their center positions $C_{t} \in C$ being within 10m from $C_{i}$. 
Next, from all of the selected rectangles, we randomly sample world positions to place human models. 
The sampled positions are at least 1m far from each other. 
Lastly, we place human models by adjusting the heights of the lowest vertices of the human models to those of the sampled positions. 
About five models are spawned on average at once, and we render egocentric views from them. 
After that, we go back to the first step.
This way, we randomly place the human models closer to each other in the 3D environment, and some rendered views can capture multiple models. 

\smallskip
\noindent\textbf{Rendering.} 
We use a fisheye plugin~\cite{ssc} to render images until motions are completed, or a collision is detected. Here, we use the physics engine of Unreal Engine to detect collisions based on the pre-defined collision proxies (volumes) of the human models and the 3D environments. 
Around 100 stereo views per motion are rendered on average.
The environments contain multiple light sources, including sky, points, and directional lights. 
Ray-tracing is enabled if the environments support it;  rasterization rendering is used otherwise. 
Also, the rendering process of Unreal Engine includes deferred shading, global illumination, lit translucency, post-processing, and GPU particle simulation utilizing vector fields. 
Please refer to our supplement for more details on the asset rendering. 
All images are rendered on NVIDIA RTX 3090. 
The rendering speed is two frames per second.

\subsection{Egocentric Dataset}

We capture stereo fisheye images and depth maps with a resolution of $1024{\times}1024$ pixels each with 25 frames per second. 
Metadata is provided for each frame, including 3D joint positions, camera positions, and 2D coordinates of reprojected joint positions in the fisheye views. 
We randomly choose $10\%$ of our motion data over all motion types, and capture 450k in-the-wild stereo views (900k images) in total. 
See Fig.~\ref{fig:egocap} for the comparison with existing egocentric datasets. 

As mentioned in Sec.~\ref{related work:egocentric data}, the motion variety is our top priority. 
UnrealEgo contains many complex motions in daily activities, some of which are difficult to capture with corresponding egocentric views in real-world settings. 
Example motions include breakdance and backflip in the dancing category shown in Table~\ref{table:motion_category_unrealego}. 
To highlight the diversity of motions in UnrealEgo, we visualize the distributions of the keypoints in our UnrealEgo and xR-EgoPose \cite{tome2019xr} datasets in Fig. \ref{fig:distribution of keypoints}. 
Here, we use pelvis-relative 3D coordinates for head and left foot positions. 
Overall, the keypoints of UnrealEgo are more widespread with a larger variance of distances from the pelvis (origin) than those of xR-EgoPose. For example, in the left 3D plot of Fig.~\ref{fig:distribution of keypoints}, the head is moving through a larger 3D space in UnrealEgo, even to areas below the pelvis, whereas head locations of xR-EgoPose are 
predominantly 
fixed above the pelvis. 
This shows that the UnrealEgo motions have 
a higher diversity of head positions.

\section{Egocentric 3D Human Pose Estimation} 
\begin{figure}[t]
 \centering
 \includegraphics[width=12cm]{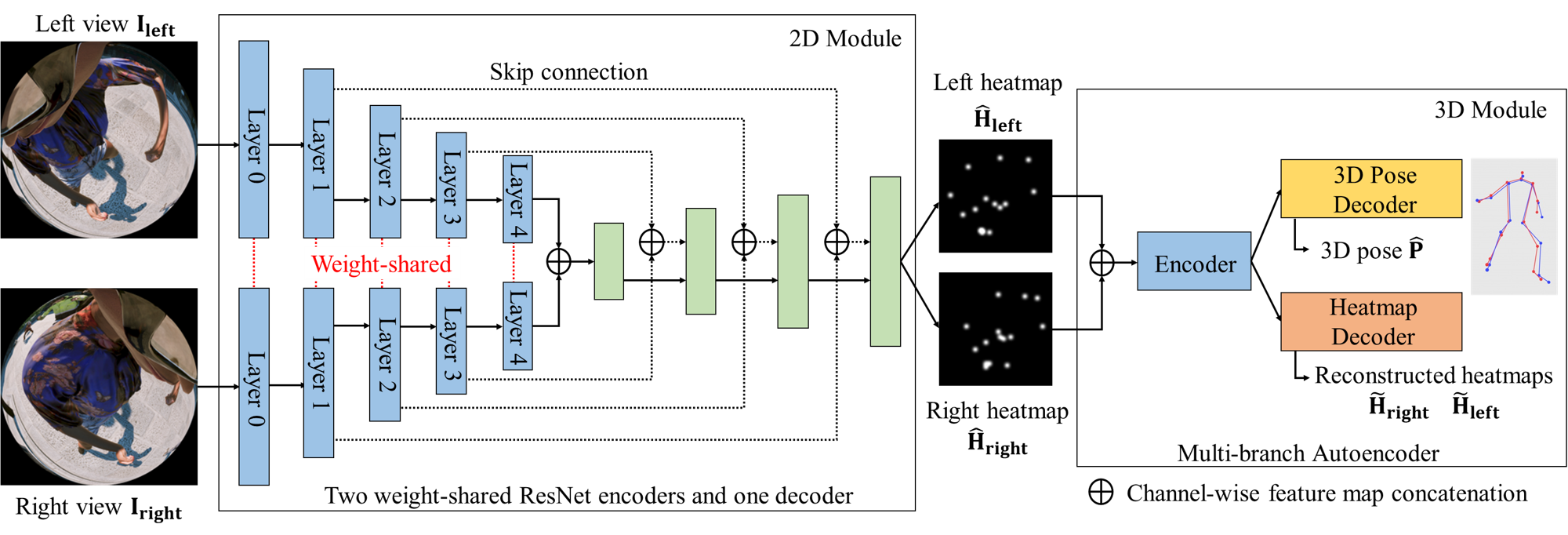}
 \caption{Overview of the proposed method. Our network consists of a 2D module to predict 2D heatmaps of joint positions from stereo inputs (Sec.~\ref{2d heatmap module}) and a 3D module to estimate 3D joint positions from the heatmaps (Sec.~\ref{3d module}).}
 \label{fig:overview of method}
\end{figure}

In this section, we describe our egocentric 3D human pose estimation method. 
We firstly adopt a 2D module to predict 2D heatmaps of joint positions from stereo inputs and, next, a 3D module to  generate 3D joint positions from the 2D heatmaps. 
Fig.~\ref{fig:overview of method} shows the overview of our network architecture. 
The main contribution of our method lies in the 2D module specifically designed for stereo inputs. 
This differs from the previous work~\cite{zhao2021egoglass},  which uses two separate 2D modules for stereo views. 
In the following, we explain each module in detail.

\subsection{2D Module}
\label{2d heatmap module}
Our 2D module consists of two weight-sharing encoders and one decoder with unified skip connections~\cite{ronneberger2015u} for stereo features as shown in Fig.~\ref{fig:overview of method}. 
Here, we follow~Zhao \textit{et  al.}~\cite{zhao2021egoglass} to use ResNet18~\cite{he2016deep} as our encoder backbone. 
The 2D module takes stereo RGB images $\{\mathbf{I}_{\text{Left}}, \mathbf{I}_{\text{Right}}\} \in \mathbb{R}^{256 \times 256 \times 3 }$ as inputs, and infers 2D joint locations represented as a set of heatmaps $\{\mathbf{H}_{\text{Left}}, \mathbf{H}_{\text{Right}}\} \in \mathbb{R}^{64 \times 64 \times 15}$. 
Here, we predict 15 joints in the neck, upper arms, lower  arms, hands, thighs, calves, feet and balls of the feet. 
From each layer of the two weight-sharing encoders, we extract the features and concatenate them along the channel dimension. 
These features are then forwarded to corresponding decoder layers via skip connections. 
Unlike the 2D module of the previous work~\cite{zhao2021egoglass}, our 2D module 
utilizes stereo information for heatmap estimation and, thus, boosts the performance of the 3D pose estimation task. 
For the training of the 2D module, we apply the mean squared error (mse) between the ground-truth heatmaps $\mathbf{H}_{\text{Left}}$ and $\mathbf{H}_{\text{Right}}$ and the estimated 2D heatmaps $\widehat{\mathbf{H}}_{\text{Left}}$ and $\widehat{\mathbf{H}}_{\text{Right}}$: 
\begin{equation}
  L_{\text{2D}} = \text{mse}(\mathbf{H}_{\text{Left}}, \widehat{\mathbf{H}}_{\text{Left}}) + \text{mse}(\mathbf{H}_{\text{Right}}, \widehat{\mathbf{H}}_{\text{Right}}).
\end{equation}

\begin{table}[t]
\begin{center}
\caption{Comparisons on UnrealEgo with and w/o  
ImageNet pre-training.} 
\label{table:quantitative evaluation} 
\scalebox{0.80}{
\begin{tabular}{lccc}
\hline
\noalign{\smallskip}
Methods    
& \,\,\,\,\,\,\,\,\,\,\,\,Settings\,\,\,\,\,\,\,\,\,\,\,\,
& \,\,\,\,\,\,\,\,\,\,\,\,\,\,\,\,\,\,\,\,\,\,\,\,\,\,\,MPJPE ($\sigma$) \,\,\,\,\,\,\,\,\,\,\,\,\,\,\,\,\,\,\,\,\, \,\,\,\,\,\,
& \,\,\,\,\,\,\,\,\,\,\,\,\,\,\,\,\,\,\,\,\,\,\,\,\,\,\,PA-MPJPE ($\sigma$)\,\,\,\,\,\,\,\,\,\,\,\,\,\,\,\,\,\,\,\,\,\,\,\,\,\,\,\\

\noalign{\smallskip}
\hline
\noalign{\smallskip}
xR-EgoPose                             & Monocular & 112.86 (1.16) / 123.15 (2.05)  & 88.71 (0.98) / 96.56 (1.27)  \\
\noalign{\smallskip}
\hdashline
\noalign{\smallskip}
EgoGlass                               & Stereo    & 91.44 (0.84) / 107.70 (1.88)    & 70.21 (0.90) / 84.22 (0.99)  \\
\bf{Ours}                              & Stereo  & \bf{79.06 (0.25) /} \bf{87.31 (0.57)}     & \bf{59.95 (0.74) /} \bf{64.65 (0.93)}  \\

\noalign{\smallskip}
\hline
\noalign{\smallskip}

\end{tabular}
}
\end{center}
\end{table}

\subsection{3D Module}
\label{3d module}
Following previous work~\cite{zhao2021egoglass}, we adopt the same multi-branch autoencoder for our 3D module. Given the heatmaps $\widehat{\mathbf{H}}_{\text{Left}}$ and $\widehat{\mathbf{H}}_{\text{Right}}$ predicted by the 2D module as inputs, the 3D module firstly encodes them to get embedding features. These features are used in two decoder branches. The first branch is a 3D pose branch, which outputs the final 3D pose ${\hat{\mathbf{P}} \in \mathbb{R}^{16\times3}}$. Here, the number of output 3D joints is 16 as the head position is included. The second branch is a heatmap branch, which tries to reconstruct the predicted 2D heatmaps ${\widetilde{\mathbf{H}}}_{\text{Left}}$ and ${\widetilde{\mathbf{H}}}_{\text{Right}}$ so that the network can capture the uncertainty of the heatmaps. 

Similar to~\cite{zhao2021egoglass}, the overall loss function for the 3D module is as follows: 
\begin{eqnarray} 
  L_{\text{3D}} & = & \lambda_{\text{pose}}(\text{mpjpe}(\mathbf{P}, \hat{\mathbf{P}}) + \lambda_{\text{cos}}\text{cos}(\mathbf{P}, \hat{\mathbf{P}})) + \nonumber \\    
  & & \lambda_{\text{hm}}(\text{mse}(\widehat{\mathbf{H}}_{\text{Left}}, \widetilde{\mathbf{H}}_{\text{Left}}) + \text{mse}(\widehat{\mathbf{H}}_{\text{Right}}, \widetilde{\mathbf{H}}_{\text{Right}})),
\end{eqnarray}
where $\mathbf{P}$ is a ground-truth 3D pose, mpjpe($\cdot$) is %
the mean per joint position error and cos($\cdot$) is a 
negative cosine similarity, \textit{i.e.,} %

\begin{equation}
  \text{mpjpe}(\mathbf{P}, \hat{\mathbf{P}}) = \frac{1}{BJ}\sum_{i=1}^{B}\sum_{j=1}^{J}||\mathbf{P}_{i}^j - \hat{\mathbf{P}}_{i}^j||_2, 
\end{equation}
\begin{equation}
  \text{cos}(\mathbf{P}, \hat{\mathbf{P}}) = -\frac{1}{B}\sum_{i=1}^{B}\sum_{l=1}^{L}\frac{\mathbf{P}_i^l \cdot \hat{\mathbf{P}}_i^l }{||\mathbf{P}_i^l|| \, ||\hat{\mathbf{P}}_i^l||},
\end{equation}
where $B$ is the batch size, $J$ is the number of joints, $L$ is the number of limbs, and $\mathbf{P}_i^l \in \mathbb{R}^3$ is the $l$-th  bone of the human skeleton.

\begin{table}[t]
\begin{center}
\caption{Quantitative evaluation on the general motions of UnrealEgo (MPJPE).}
\label{table:quantitative evaluation general}
\scalebox{0.69}{
\begin{tabular}{lcccccccc}
\hline
\noalign{\smallskip}
Methods                                
& jumping  
& falling down  
& exercising  
& pulling, 
& singing  
& rolling 
& crawling 
& laying\\

\noalign{\smallskip}
\hline
\noalign{\smallskip}
xR-EgoPose          & 106.30      & 167.18   & 133.19 & 119.49 & 99.62 & 166.14  & 223.51     & 146.67 \\
\noalign{\smallskip}
\hdashline
\noalign{\smallskip}
EgoGlass            & 88.55       & 135.25   & 105.11 & 89.96  & 75.54 & 143.64  & 199.27     & 114.85\\
\bf{Ours}                & \bf{76.81}  & \bf{125.22} & \bf{90.54}  & \bf{80.61}  & \bf{65.53}  & \bf{94.97}  & \bf{179.98}     & \bf{97.56}\\

\noalign{\smallskip}
\hline
\noalign{\smallskip}
\hline
\noalign{\smallskip}
Methods        

& \begin{tabular}[c]{@{}c@{}}sitting on\\ the ground \end{tabular} \, 
& \begin{tabular}[c]{@{}c@{}}crouching \\ - normal \end{tabular} \, 
& \begin{tabular}[c]{@{}c@{}}crouching \\ - turning \end{tabular} \,
& \begin{tabular}[c]{@{}c@{}}crouching \\ - to standing \end{tabular} \,
& \begin{tabular}[c]{@{}c@{}}crouching \\ - forward \end{tabular} \,
& \begin{tabular}[c]{@{}c@{}}crouching \\ - backward \end{tabular} \, 
& \begin{tabular}[c]{@{}c@{}}crouching \\ - sideways \end{tabular} \, 
& \begin{tabular}[c]{@{}c@{}}standing \\ - whole body \end{tabular} \\

\noalign{\smallskip}
\hline
\noalign{\smallskip}
xR-EgoPose              & 274.99 & 172.25 & 173.77 & 108.96   & 119.95     & 136.52   & 145.81 & 94.34 \\
\noalign{\smallskip}
\hdashline
\noalign{\smallskip}
EgoGlass                & 216.52 & 129.72 & 151.71 & 93.71  & 90.76      & 100.39   & 122.23 & 78.57\\
\bf{Ours}                   & \bf{195.28} & \bf{120.65}  & \bf{131.82} & \bf{81.28} & \bf{76.04}      & \bf{81.31}    & \bf{88.54}  & \bf{67.67} \\

\noalign{\smallskip}
\hline
\noalign{\smallskip}
\hline
\noalign{\smallskip}
Methods
 
& \begin{tabular}[c]{@{}c@{}}standing \\ - upper body \end{tabular} \,
& \begin{tabular}[c]{@{}c@{}}standing \\ - turning \end{tabular} \,
& \begin{tabular}[c]{@{}c@{}}standing \\ - to crouching \end{tabular} \,
& \begin{tabular}[c]{@{}c@{}}standing \\ - forward \end{tabular} \,  
& \begin{tabular}[c]{@{}c@{}}standing \\ - backward \end{tabular} \,
& \begin{tabular}[c]{@{}c@{}}standing \\ - sideways \end{tabular} \, 
& \,
& all \\

\noalign{\smallskip}
\hline
\noalign{\smallskip}
xR-EgoPose           & 93.36 & 103.28  & 101.60     & 99.72   & 105.86 & 114.28 &  & 112.61 \\
\noalign{\smallskip}
\hdashline
\noalign{\smallskip}
EgoGlass              & 76.83 & 84.12   & 82.03      & 82.96   & 85.15  & 93.61  &  & 91.27\\
\bf{Ours}                  & \bf{65.92} & \bf{74.55} & \bf{73.21}      & \bf{70.86}   & \bf{70.40}  & \bf{79.06}  &  & \bf{79.57} \\

\noalign{\smallskip}
\hline
\noalign{\smallskip}

\end{tabular}
}
\end{center}
\end{table}

\begin{table}[t]
\vspace{-20pt}
\begin{center}
\caption{Quantitative evaluation on the sports motions of UnrealEgo (MPJPE).}
\label{table:quantitative evaluation sport}
\scalebox{0.70}{
\begin{tabular}{lcccccccccc}
\hline
\noalign{\smallskip}
Methods                                
& \,\,\,dancing  \,\,\,
& \,\,\,boxing  \,\,\,
& \,\,\,wrestling  \,\,\,
& \,\,\,soccer \,\,\,
& \,\,\,baseball \,\,\,
& \,\,\,basketball \,\,\,
& \,\,\,american football\,\,\,
& \,\,\,golf \,\,\,
& \,\,\,all \,\,\,\\

\noalign{\smallskip}
\hline
\noalign{\smallskip}
xR-EgoPose          & 116.75     & 97.33   & 116.65 & 104.65 & 103.75  & 98.65      & 149.76   & 117.50 & 113.28   \\
\noalign{\smallskip}
\hdashline
\noalign{\smallskip}
EgoGlass            & 95.37      & 77.66   & 96.63 & 88.30 & 93.60  & 74.31      & 118.34   & 79.35  & 91.71  \\
\bf{Ours}           & \bf{79.86} & \bf{69.34} & \bf{84.02} & \bf{76.54} & \bf{74.27} & \bf{62.09}      & \bf{103.79}   & \bf{72.06} & \bf{78.19} \\

\noalign{\smallskip}
\hline
\noalign{\smallskip}

\end{tabular}
}
\end{center}
\end{table}

\section{Experiments}
\label{experiments}

\subsection{Implementation Details} 
We randomly split UnrealEgo into 3,821 motions (357,317 stereo views) for training, 494 motions (46,207 stereo views) for validation, and 526 motions (48,080 stereo views) for testing.
The input images and ground-truth 2D heatmaps are resized to  $256{\times}256$ and $64{\times}64$, respectively. 
The 2D module and the 3D module are trained separately on a Quadro RTX 8000 with a batch size of 16. 
We set the hyper-parameters as $\lambda_{\text{pose}}=0.1$, $\lambda_{\text{cos}}=0.01$, and $\lambda_{\text{hm}}=0.001$. 
The modules are trained with Adam optimizer  \cite{Kingma2015} for ten epochs, starting with a learning rate of $10^{-3}$ for the first half epochs and applying a linearly decaying rate for the next half. 
We perform the experiments three times and report average scores and standard deviations (denoted by $\sigma$). 

\subsection{Comparisons}
As our comparison methods, we adopt state-of-the-art methods for egocentric 3D human pose estimation, \textit{i.e.,} EgoGlass~\cite{zhao2021egoglass} and xR-EgoPose~\cite{tome2019xr}. 
Since their source codes are not available, 
we re-implement and tailor them for UnrealEgo. 
We train xR-EgoPose on the left views of UnrealEgo. 
For the sake of evaluation under the same conditions, we remove a body part branch with segmentation supervision in EgoGlass as xR-EgoPose does not use it.
We follow the previous works and report the Mean Per Joint Position Error (MPJPE) and the Mean Per Joint Position Error with Procrustes alignment~\cite{10.1214/ss/1177012582} (PA-MPJPE). 
Here, Procrustes alignment finds optimal rigid transformation and scale between the predicted and ground-truth 3D poses.

\begin{figure}[t]
  \centering
    \caption*{\scriptsize
        \,\,
        Stereo inputs 
        \,\,\,\,\,\,\,\,\,
        EgoGlass~\cite{zhao2021egoglass} 
        \,\,\,
        Ours
        \,\,\,\,\,\,\,\,\,\,\,\,\,\,\,\,\,\,\,\,\,\,\,\,
        Stereo inputs 
        \,\,\,\,\,\,\,\,\,
        EgoGlass~\cite{zhao2021egoglass} 
        \,\,\,
        Ours
    }
  \vspace{-5pt}
  \includegraphics[width=12.1cm]{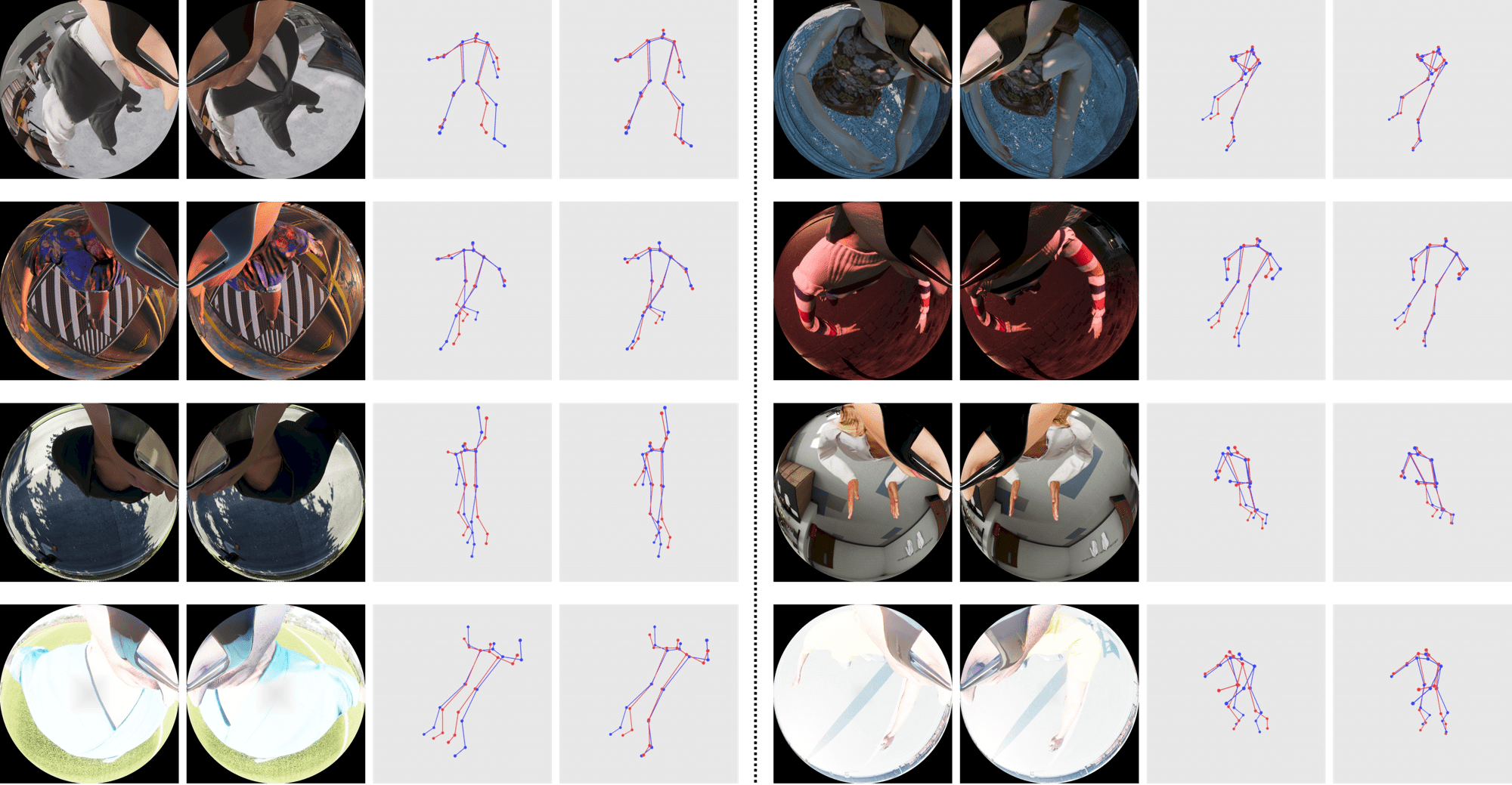}\\
  \caption{Qualitative results on UnrealEgo (blue: ground truth; red: prediction).} 
  \label{fig:qualitative result seccessful}
\end{figure}

\begin{figure}[t]
  \begin{minipage}[t]{.49\textwidth}
    \begin{center}

    \caption*{\scriptsize
        \,\,\,\,\,\,\,\,
        Stereo inputs 
        \,\,\,\,\,\,\,\,\,\,
        EgoGlass~\cite{zhao2021egoglass} 
        \,\,\,
        Ours
        \,\,
    }
  \vspace{-5pt}
  \includegraphics[width=1\textwidth]{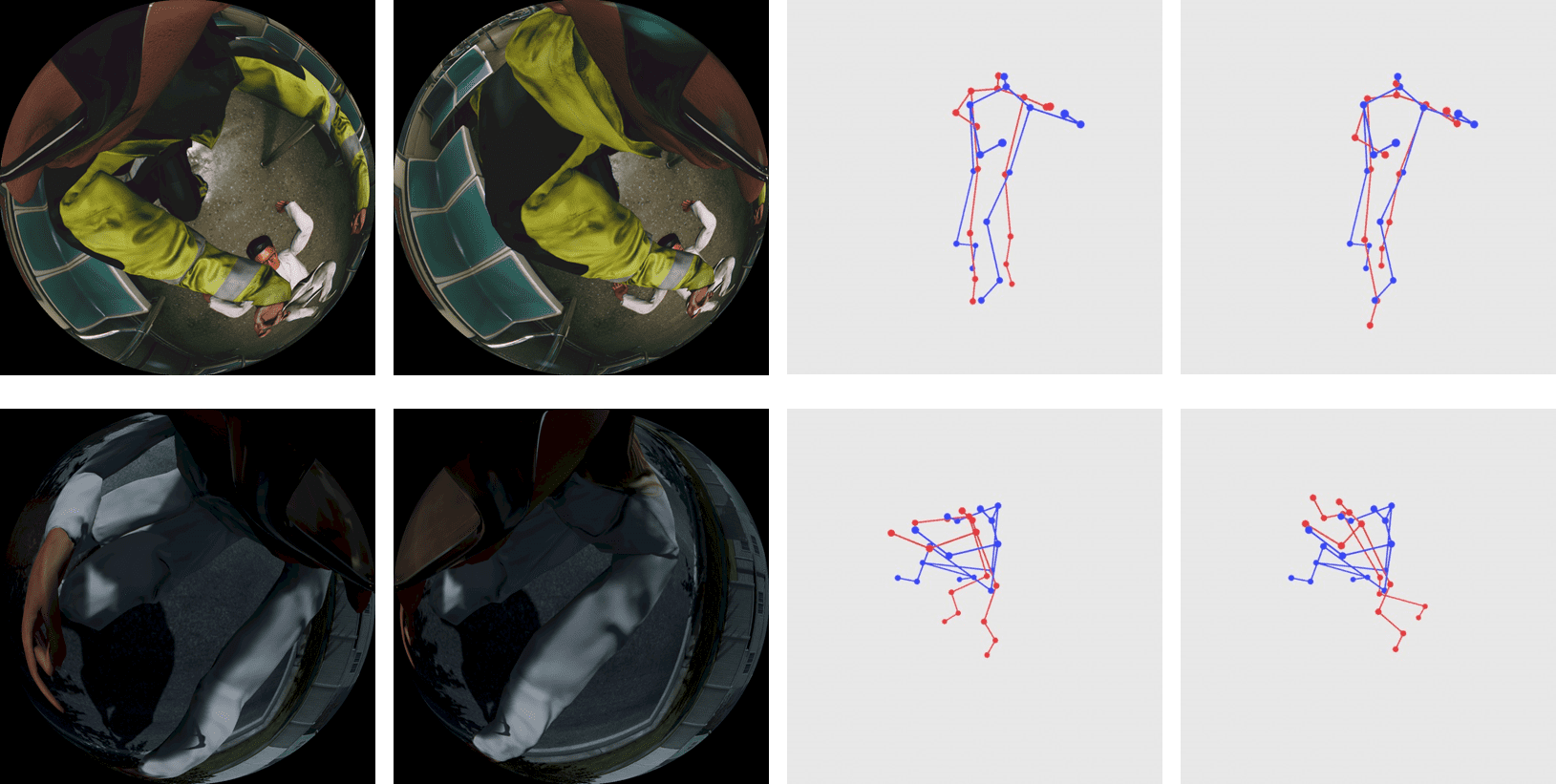}\\
  \caption{Qualitative results for failure cases on UnrealEgo.}
  \label{fig:qualitative result failure}

    \end{center}
  \end{minipage}
  \hfill
  \begin{minipage}[t]{.49\textwidth}
    \begin{center}

      \caption*{\scriptsize 
        \begin{tabular}[c]{@{}c@{}}Right view \end{tabular}
        \,
        \begin{tabular}[c]{@{}c@{}}End-to-end \end{tabular}
        \,\,\,
        \begin{tabular}[c]{@{}c@{}}Separate \end{tabular}
        \,\,\,\,\,\,\,\,\,\,\,\,
        \begin{tabular}[c]{@{}c@{}}GT\end{tabular}
        \,\,\,\,\,\,\,\,\,\,
    }
  \vspace{-5pt}
  \includegraphics[width=1\textwidth]{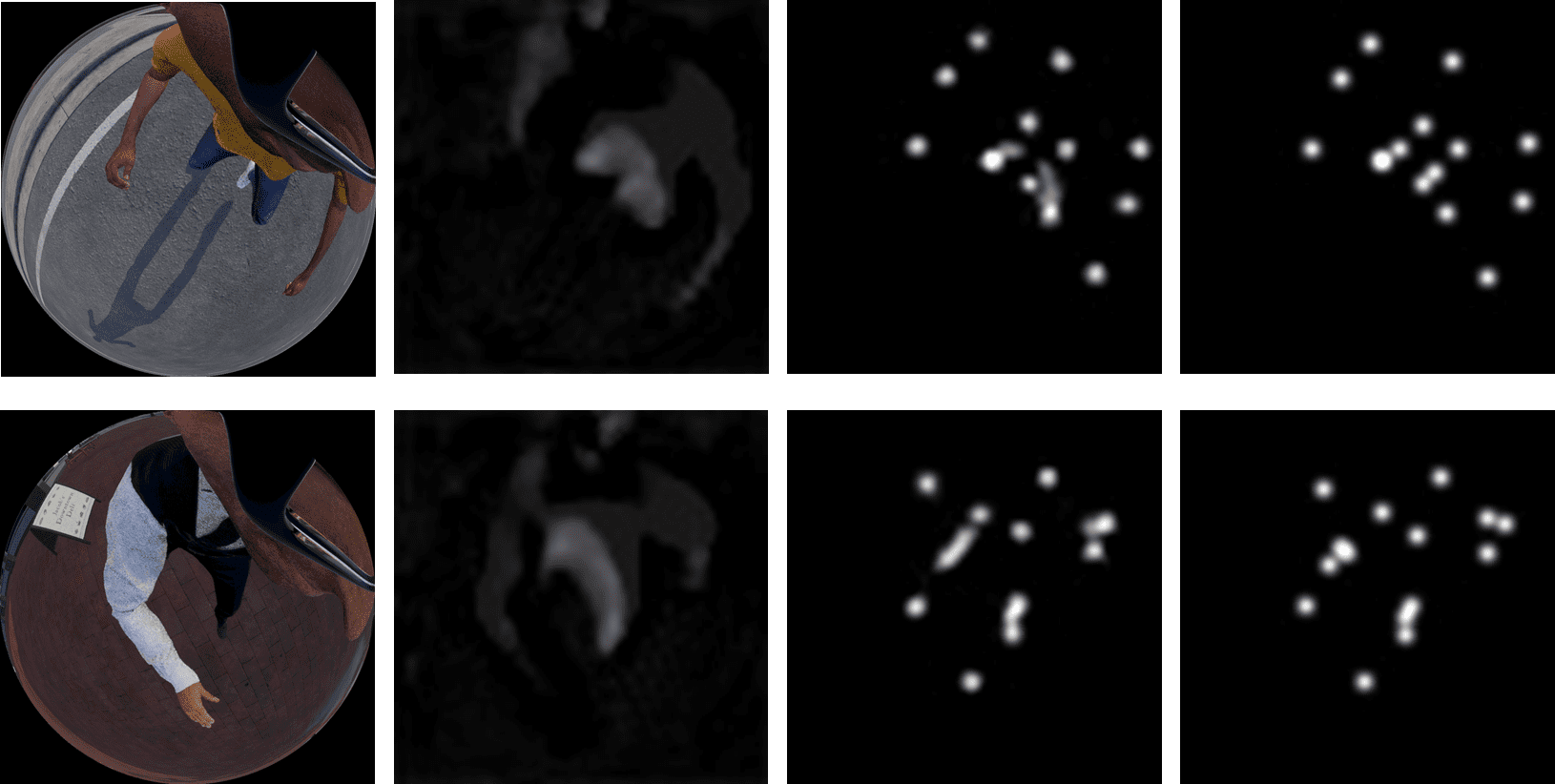}\\
  \caption{Heatmap estimation results with two different training strategies.}
  \label{fig:ablation study training strategy}

    \end{center}
  \end{minipage}
\end{figure}

\subsection{Results}
We present results on the UnrealEgo test sequence.
Table~\ref{table:quantitative evaluation} quantitatively evaluates our approach and competing methods 
with and without ImageNet pre-training for the encoder. 
Overall, our method outperforms the previous best-performing method~\cite{zhao2021egoglass}, across all metrics for both experiments with and without ImageNet. 
Specifically, our method with the pre-trained encoder shows significant improvement by 13.5\% on MPJPE and 14.65\% on PA-MPJPE compared to EgoGlass \cite{zhao2021egoglass}. 
All methods, including ours, benefit from the ImageNet pre-training; the performance of our approach is boosted by 9.4\% on MPJPE and 7.2\% on PA-MPJPE.

We also break down the test sequence into 30 motion types as shown in Table~\ref{table:quantitative evaluation general} for general motions and  Table~\ref{table:quantitative evaluation sport} for sports motions. 
Both tables indicate that our method achieves significant superiority for all motion types. 
See Fig.~\ref{fig:qualitative result seccessful} for the qualitative results. Even with the occlusions and complex poses in various environments, our method estimates the 3D poses much better than EgoGlass. 

It is also worth analyzing failure cases. 
According to Table~\ref{table:quantitative evaluation general}, 
bending motions (such as sitting on the ground or crouching) are reconstructed with comparably low accuracy. 
This is because the lower body parts are occluded by the upper body, especially when people crouch down as shown in Fig.~\ref{fig:qualitative result failure}. 
Even with the stereo inputs, these methods still can not perform well on some motions that are occasionally seen in daily human activities.

\begin{table}[t]
  \begin{minipage}[t]{.47\textwidth}
    \begin{center}
        \caption{Ablation study for the backbone of the 2D heatmap module.}
        \label{table:ablation study backbone}
        \scalebox{0.80}{
        \begin{tabular}{lcc}
        \hline
        \noalign{\smallskip}
        Backbones\,\,    
        & MPJPE ($\sigma$) \, 
        & PA-MPJPE ($\sigma$)\\
        
        \noalign{\smallskip}
        \hline
        \noalign{\smallskip}
        ResNet18                             & \bf{79.06 (0.25)}   & \bf{59.95 (0.74)}  \\
        ResNet34                             & 80.50 (0.78)   & 60.04 (0.60)  \\
        ResNet50                             & 80.07 (0.45)   & 60.08 (0.63)  \\
        ResNet101                            & 80.15 (0.06)   & 60.57 (0.79)  \\
        
        \noalign{\smallskip}
        \hline
        \noalign{\smallskip}
      \end{tabular}
      }
    \end{center}
  \end{minipage}
  \hfill
  \begin{minipage}[t]{.50\textwidth}
    \begin{center}

        \caption{Ablation study for the weight sharing in the 2D heatmap module.}
        \label{table:ablation study weight sharing}
        \scalebox{0.80}{
        \begin{tabular}{lcc}
        \hline
        \noalign{\smallskip}
        Backbones\,\,    
        & MPJPE ($\sigma$) \, 
        & PA-MPJPE ($\sigma$)\\
        
        \noalign{\smallskip}
        \hline
        \noalign{\smallskip}
        weight sharing            & \bf{79.06 (0.25)}   & \bf{59.95 (0.74)}  \\
        no weight sharing         & 83.54 (1.30)   & 62.29 (0.45)  \\
        
        \noalign{\smallskip}
        \hline
        \noalign{\smallskip}
      \end{tabular}
      }

    \end{center}
  \end{minipage}
\end{table}

\subsection{Ablation Study} 
We first ablate different encoder backbone architectures for our 2D module in Table~\ref{table:ablation study backbone}. All variants generate the heatmap with the same resolution and the 3D module shares the same architecture. The experiment suggests that all of the models yield similar results but at a higher computational cost for a larger backbone. For example, the difference between ResNet18 and Resnet50 is only 0.2\% on PA-MPJPE. This result is also observed in the previous work~\cite{Tom2020SelfPose3E}, showing that a larger backbone does not necessarily lead to performance improvements.

Next, we show the effect of weight sharing in the encoder backbone of our 2D keypoint estimation module in Table~\ref{table:ablation study weight sharing}. The weight-sharing backbone performs better than the encoder without weight sharing by 5.4\% on MPJPE and 3.8\% on PA-MPJPE. One possible reason for this result is that the weight-sharing backbone can see more views during training, leading to a better feature extractor. Therefore, we use the weight-sharing strategy for all experiments.

\begin{table}[t]
\begin{center}
\caption{Ablation study on the training strategy.}
\label{table:ablation study training strategy}
\scalebox{0.80}{
\begin{tabular}{lcc}
\hline
\noalign{\smallskip}
Backbones\,\,    
& MPJPE ($\sigma$) \, 
& PA-MPJPE ($\sigma$)\\

\noalign{\smallskip}
\hline
\noalign{\smallskip}
Separate training            & \bf{79.06 (0.25)}   & \bf{59.95 (0.74)}  \\
End-to-end training         & 80.67 (0.58)   & 61.72 (0.55)  \\

\noalign{\smallskip}
\hline
\noalign{\smallskip}

\end{tabular}
}
\end{center}
\end{table}

Lastly, we conduct the experiment with different training strategies, \ie separate training and end-to-end training for our 2D keypoint estimation and 3D estimation module, as shown in Table~\ref{table:ablation study training strategy}. The result indicates that the separate training yields slightly better performance than the end-to-end training by 2.0\% on MPJPE and 2.9\% on PA-MPJPE. We also visualize the heatmaps predicted by our network with the different training strategies in Fig.~\ref{fig:ablation study training strategy}. It is interesting to note that separate training leads to relatively accurate heatmap estimation while the network trained in an end-to-end manner tries to capture the whole body. Although this visual result can  change depending on the hyper-parameters, we follow the same hyper-parameter setting in the previous work~\cite{zhao2021egoglass} and choose the separate training strategy for all experiments.

\section{Conclusions} 

We presented \textit{UnrealEgo}, \textit{i.e.,} a new large-scale naturalistic dataset for egocentric 3D human pose estimation. 
It allows a comprehensive evaluation of existing and upcoming methods for egocentric 3D vision, including the temporal component and global 3D poses. 
Our simple yet effective architecture for egocentric 3D human pose estimation brings significant improvement compared to previous best-performing methods qualitatively and quantitatively. 
In addition, our extensive ablation studies validate our architectural design choices for the stereo inputs and the training strategy. 
Although our method achieved state-of-the-art results, there are still failure cases due to occlusions and complex motions.
In future work, we are interested in incorporating explicit 3D geometry obtained from our stereo fisheye setup for further performance improvements. 

{\small
\subsubsection{Acknowledgements.} 
We thank Silicon Studio Corp. for providing the fisheye plugin. 
Hiroyasu Akada and Masaki Takahashi were supported by the Core Research for Evolutional Science and Technology of the Japan Science and Technology Agency (JPMJCR19A1). 
Jian Wang, Soshi Shimada, Vladislav Golyanik and Christian Theobalt were supported by the ERC Consolidator Grant 4DReply (770784). 
}

\clearpage
\bibliographystyle{splncs04}
\bibliography{egbib}

\setcounter{section}{0}
\renewcommand\thesection{\Alph{section}}
\newcommand{\suppsection}{\subsection}
\clearpage

\title{UnrealEgo: A New Dataset for Robust Egocentric 3D Human Motion Capture} 
\subtitle{Supplementary Material}

\titlerunning{UnrealEgo: A New Dataset for Robust Egocentric 3D Human MoCap}

\author{Hiroyasu Akada\inst{1,2}
\and
Jian Wang\inst{1} 
\and
Soshi Shimada\inst{1} \and
Masaki Takahashi\inst{2} \and \\
Christian Theobalt\inst{1} \and
Vladislav Golyanik\inst{1}
}
\authorrunning{H. Akada et al.}
\institute{
Max Planck Institute for Informatics, SIC 
\and
Keio University 
}

\makeatletter

\newpage
 \markboth{}{}%
 \def\lastand{\ifnum\value{@inst}=2\relax
                 \unskip{} \andname\
              \else
                 \unskip \lastandname\
              \fi}%
 \def\and{\stepcounter{@auth}\relax
          \ifnum\value{@auth}=\value{@inst}%
             \lastand
          \else
             \unskip,
          \fi}%
 \begin{center}%
 \let\newline\\
 {\Large \bfseries\boldmath
  \pretolerance=10000
  \@title \par}\vskip .8cm
\if!\@subtitle!\else {\large \bfseries\boldmath
  \vskip -.65cm
  \pretolerance=10000
  \@subtitle \par}\vskip .8cm\fi
 \setbox0=\vbox{\setcounter{@auth}{1}\def\and{\stepcounter{@auth}}%
 \def\thanks{}\@author}%
 \global\value{@inst}=\value{@auth}%
 \global\value{auco}=\value{@auth}%
 \setcounter{@auth}{1}%
{\lineskip .5em
\noindent\ignorespaces
\@author\vskip.35cm}
 {\small\institutename}
 \end{center}%
\makeatother

This supplementary material provides more details on our assets, including realistic 3D human models and  environments, and motion modification. Please also watch the supplementary video\footnote{\url{https://4dqv.mpi-inf.mpg.de/UnrealEgo/}} for dynamic visualization of UnrealEgo.

\def\thesection{\Alph{section}}
\setcounter{section}{0}
\setcounter{figure}{8}
\setcounter{table}{8}
\setcounter{equation}{4}

\section{Asset List}

\subsection{Characters}
\begin{figure}[H]
  \centering
  \includegraphics[width=8.5cm]{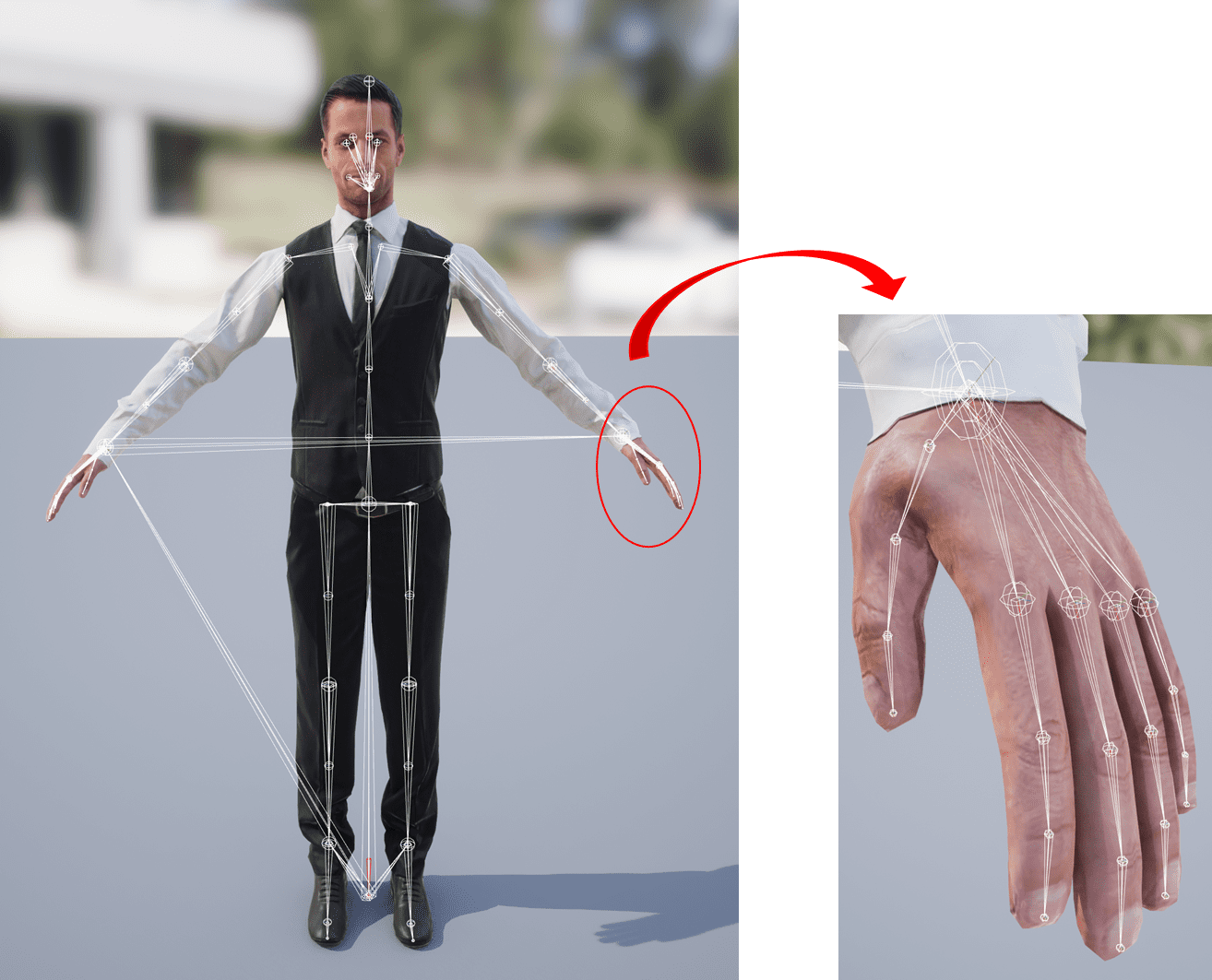}\\
  \caption{The skeleton of the RenderPeople models.} 
  \label{fig:skeleton} 
\end{figure}

We use 17 realistic RenderPeople 3D human models (commercially available) \cite{renderpeople}, \textit{i.e.,} nine female and eight male. 
Table~\ref{table:list of characters} summarizes the RenderPeople models that we use for UnrealEgo. 
These models are rigged and skinned based on the default 3D human skeleton of Unreal Engine~\cite{unrealengine} as shown in Fig.~\ref{fig:skeleton}. 
Readers are referred to RenderPeople web page \cite{renderpeople} for more details on their human models. 

\begin{table}[H]
\begin{center}
\caption{List of characters.} 
\label{table:list of characters}
\scalebox{0.85}{
\begin{tabular}{lllll}
\hline
Model name  \,\,\,\,\,\,\,\,\,\,\,\,\,\,\,\,\,\,\,\,\,\,\,\,
& Gender \,
& Skin color \,\,\,\,\,\,\,\,\,\,\,\,\,\,\,\,\,
& Clothes\,\,\,\,\,\,\,\,\,\,\,\,\,\,\,\,\,\,\,\,\,\,\,\,\,\,\,\,\,\,\,\,\,\,\,\,\,\,\,\,\,\,\,\,\,\,\,\,\,\,\,\,\,\,\,\,\,\,\,\,\,\,\,\,\,\,\,\,\,\,\,\,\,
& Shoes  \\
\hline
rp\_adanna\_rigged\_003    & Female  & Dark brown       & Scrubs   & Sneakers \\
rp\_amit\_rigged\_003      & Male    & Dark brown       & T-shirts, jeans  & Sneakers       \\
rp\_carla\_rigged\_001     & Female  & Dark brown       & Suit jackets, slacks  &  High heels      \\
rp\_claudia\_rigged\_002   & Female  & Pale white       & Long sleeves, slacks  & High heels     \\
rp\_eric\_rigged\_001      & Male    & Light brown      & Long sleeves, vest, slacks & Brogues      \\
rp\_janna\_rigged\_002     & Female  & White            & Scrubs & Sneakers     \\
rp\_joko\_rigged\_003      & Male    & White            & Turnout coat, bunker gear  &  Bunker gear     \\
rp\_joyce\_rigged\_005     & Female  & Black            & Dress with long pants  & High heels      \\
rp\_kyle\_rigged\_001      & Male    & Light brown      & Scrubs   & Sneakers      \\
rp\_manuel\_rigged\_001    & Male    & White            & T-shirts, jeans  & Sneakers      \\
rp\_maya\_rigged\_003      & Female  & Pale white       & Long sleeves, shorts, tights   &  Sneakers     \\
rp\_nathan\_rigged\_003    & Male    & Light brown      & T-shorts, jeans  & Sneakers      \\
rp\_rin\_rigged\_007       & Female  & White            & Dress  & High heels      \\
rp\_scott\_rigged\_005     & Male    & White            & T-shorts, safety vest, athletic pants & Hiking boots     \\
rp\_serena\_rigged\_004    & Female  & Black            & Short sleeves, shorts  & Sneakers      \\
rp\_shawn\_rigged\_004     & Male    & Moderate brown   & Scrubs  & Sneakers      \\
rp\_sophia\_rigged\_003    & Female  & Moderate brown   & No sleeves, jeans  & Flats      \\

\hline
\end{tabular}
}
\end{center}
\end{table}

\subsection{3D Environments}

We provide the asset list of 3D environments used for UnrealEgo in Table~\ref{table:list of 3d environments}. All of the environments are commercially available on the UnrealEngine marketplace.
Ray tracing is enabled if the environments support it or rasterization rendering is used otherwise. Also, the rendering process of UnrealEngine includes deferred shading, global illumination, lit translucency, post-processing, and GPU particle simulation utilizing vector fields. Please also take a look at the UnrealEngine documentation~\cite{unrealenginerendering} for more details on their rendering system.

\begin{table}[H]
\begin{center}
\caption{List of 3D environments used in UnrealEgo.} 
\label{table:list of 3d environments}
\scalebox{0.76}{
\begin{tabular}{lcl}
\hline
Environment name   \,\,\,\,\,\,\,\,\,\,\,\,\,\,\,\,\,\,\,\,\,\,\,\,\,\,\,\,\,\,\,\,\,\,\,\,\,\,\,\,\,\,\,\,\,\,\,\,\,\,\,\,\,\,\,\,\,\,\,\,\,\,\,\,\,\,\,\,\,\,\,\,
& Ray-tracing 
& Example scenes \\
\hline
ArchViz Interior~\cite{archvizinterior}  & \checkmark & North-American rooms  \\
Big Office~\cite{bigoffice}   & \checkmark & Offices, cafeterias, playrooms, restrooms, elevators\\
Downtown West Modular Pack~\cite{downtownwest} & \checkmark &  Outdoor shopping mall, roads, water fountain \\
Hutong/Chinese Alleyway Pack~\cite{hutong} & \checkmark & Alleys, bicycle parking  \\
Japanese Restaurant Interior \& Exterior~\cite{japaneserestaurant} & \checkmark & Traditional Japanese izakaya  \\
Realistic Lab. Laboratory Equipment~\cite{lab} & \checkmark & Lab rooms \\
Modern Chinese Interior~\cite{modernchinese}  & \checkmark &  Chinese rooms \\
Modular Building Set~\cite{modularbuilding}  & \checkmark & Old buildings \\
Kyoto Alley~\cite{kyoto} &  &  Traditional Kyoto shopping roads, temples \\
Science Lab~\cite{sciencelab}   &  & Lab rooms  \\
City Subway Train Modular~\cite{subwaytrain}  &  & Trains \\
CityPark~\cite{citypark}  &  & \begin{tabular}[l]{@{}l@{}}Parks, roads, bridges, gardens, tennis courts, \\baseball fields, football fields, water fountain  \end{tabular} \\
Factory Environment Collection~\cite{factory} &  & \begin{tabular}[l]{@{}l@{}}Heavy truck manufacturing lines, \\warehouses, offices, changing rooms  \end{tabular} \\
Suburb Neighborhood House Pack~\cite{suburb}  &  & \begin{tabular}[l]{@{}l@{}}North-American houses, kitchens, rooms, \\stairs, gardens, pools, roads \end{tabular}\\

\hline
\end{tabular}
}
\end{center}
\end{table}

\section{Motions}

\begin{figure*}[t]
  \centering
  \includegraphics[width=12cm]{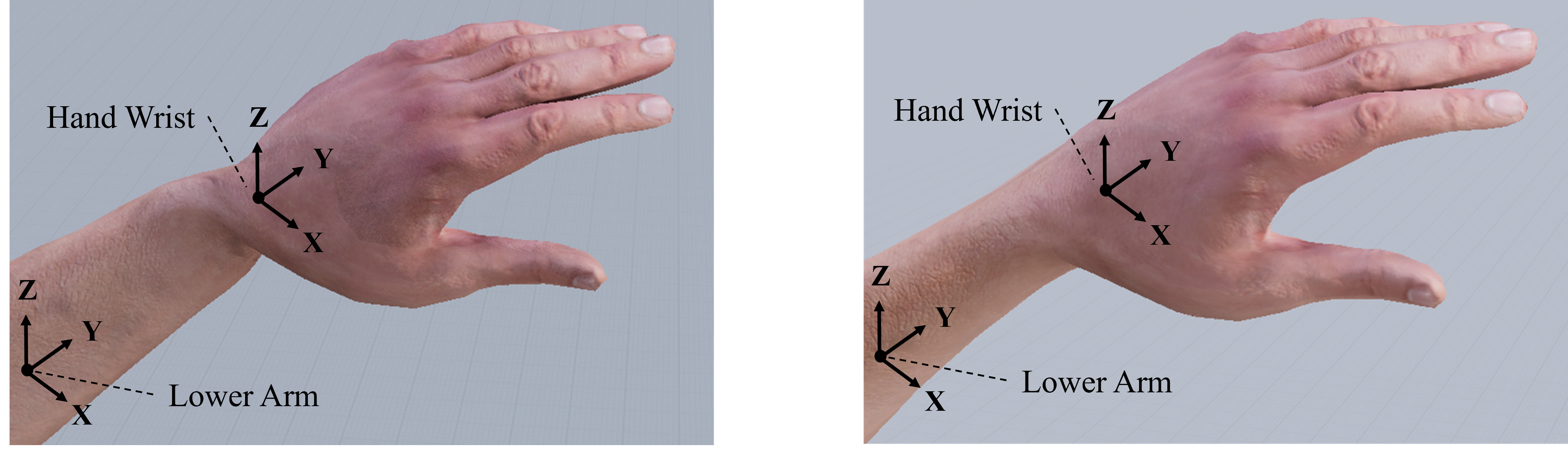}\\
    \caption*{\scriptsize \,
        (a) Original lower arm
        \,\,\,\,\,\,\,\,\,\,\,\,\,\,\,\,\,\,\,\,\,\,\,
        \,\,\,\,\,\,\,\,\,\,\,\,\,\,\,\,\,\,\,\,\,\,\,
        \,\,\,\,\,\,\,\,\,\,\,\,\,\,\,\,\,\,\,\,\,\,\,
        (b) Modified lower arm 
    }
  \caption{Modification for the lower arm distortion in Mixamo motions~\cite{mixamo}. The lower arms of the original Mixamo motions often show unnatural distortion. Therefore, we provide 70$\%$ of rotation of hand wrists around the Y axis to that of lower arms to maximize the plausibility of arm movements.} 
  \label{fig:motions} 
\end{figure*}

As mentioned in Section 3.1 of the main paper, we utilize Mixamo motions~\cite{mixamo} and modify them using UnrealEngine \cite{unrealengine} to enhance their plausibility and diversify the motion data.
Here, we use the default functions of UnrealEngine, \ie the control rig, to manually fix some motions that involve self-penetration or unnatural body distortion. In particular, we work on the problem of lower arm distortion. 
The main issue here is that the Mixamo motions lack rotation information of the lower arms of the UnrealEngine skeleton. Due to this, the original motions show unnatural distortion of the lower arms as shown in Fig.~\ref{fig:motions}-(a). To alleviate this issue, we add constraints on the amount of rotations of the lower arms. Specifically, we provide 70$\%$ of rotations of hand wrists around the Y axis to those of the lower arms to maximize the plausibility of arm movements. We show the modified version in Fig.~\ref{fig:motions}-(b). 

Also, we manually diversify the motions in various ways, including the speed of motions, arm movements, foot stance, and head rotations. The change of the head rotations is especially important for egocentric datasets because the slight change will lead to a dramatic change of egocentric views even with similar poses. Please note that xR-EgoPose~\cite{tome2019xr} also uses the Mixamo motions as mentioned in Section 2.2 (main paper). However, our manual modifications allow UnrealEgo to provide motions with more different types of poses than xR-EgoPose as discussed in Section 3.2 (main paper). 

Moreover, unlike previously proposed datasets~\cite{tome2019xr,xu2019mo2cap2}, UnrealEgo does not contain the exact same motions captured in multiple different scenes. This makes UnrealEgo a unique dataset with the largest variety of motions.

\end{document}